\documentclass[letterpaper]{article} 
\usepackage[submission]{aaai24}
\usepackage{amsmath}
\usepackage{times}  
\usepackage{authblk}
\usepackage{helvet}  
\usepackage{subfigure}
\usepackage{courier}  
\usepackage[hyphens]{url}  
\usepackage{graphicx} 
\usepackage{natbib}  
\usepackage{xcolor}
\usepackage{caption} 
\usepackage{diagbox}
\usepackage{amsfonts}
\usepackage{algorithm}
\usepackage{algorithmic}

\usepackage{newfloat}
\usepackage[marginal]{footmisc}

\usepackage{listings}
\DeclareCaptionStyle{ruled}{labelfont=normalfont,labelsep=colon,strut=off} 
\floatstyle{ruled}
\newfloat{listing}{tb}{lst}{}
\floatname{listing}{Listing}
\title{Transform then Explore: a Simple and Effective Technique for Exploratory Combinatorial Optimization with Reinforcement Learning}

\author[1,*]{Tianle Pu}
\author[1,*]{Changjun Fan}
\author[2,*]{Mutian Shen}
\author[3,4,*]{Yizhou Lu}
\author[1]{Li Zeng}
\author[2]{Zohar Nussinov}
\author[1]{Chao Chen}
\author[1]{Zhong Liu}

\affil[1]{College of Systems Engineering, National University of Defense Technology, China}
\affil[2]{Department of Physics, Washington University in St. Louis, Campus Box 1105, 1 Brookings Drive, St. Louis, MO 63130, USA}
\affil[3]{Huaxia Bank, Beijing 100005, China}
\affil[4]{Big Data and Risk Management Research Center, University of International Business and Economics, Beijing 100029, China}

\date{}

\begin{document}

\newcommand{\FC}[1]{{\bf\color{red}[{\sc FC:} #1]}}
\newcommand{\MS}[1]{{\color{blue}{#1} }}
\newcommand{\stitle}[1]{\vspace{1ex} \noindent{\bf #1}}

\maketitle
\footnote{* These authors contribute equally to this work. Corresponding author: Changjun Fan and Chao Chen.}
\begin{abstract}
Many complex problems encountered in both production and daily life can be conceptualized as combinatorial optimization problems (COPs) over graphs. Recent years, reinforcement learning (RL) based models have emerged as a promising direction, which treat the COPs solving as a heuristic learning problem. However, current finite-horizon-MDP based RL models have inherent limitations. They are not allowed to explore adquately  for improving solutions at test time, which may be necessary given the complexity of NP-hard optimization tasks. Some recent attempts solve this issue by focusing on reward design and state feature engineering, which are tedious and ad-hoc. In this work, we instead propose a much simpler but more effective technique, named gauge transformation (GT). The technique is originated from physics, but is very effective in enabling RL agents to explore to continuously improve the solutions during test. Morever, GT is very simple, which can be implemented with less than 10 lines of Python codes, and can be applied to a vast majority of  RL models. Experimentally, we show that traditional RL models with GT technique produce the state-of-the-art performances on the MaxCut problem. Furthermore, since GT  is independent of any RL models, it can be seamlessly integrated into various RL frameworks, paving the way of these models for more effective explorations in the solving of general COPs. 
\end{abstract}
\section{Introduction}
Combinatorial optimization problems(COPs) are ubiquitous in real-world applications, ranging from industry\cite{grotschel1991}, transportation\cite{laporte1992} to military planning\cite{peng2013}.  Addressing these problems efficiently has profound practical implications. Nonetheless, the intrinsic NP-hard\cite{karp1972} nature of many of these challenges makes them difficulty to solve. Existing methods often face a dilemma: fast algorithms fail to ensure solution quality, whereas solutions of higher quality often takes longer time. As a matured research topic for several decades, three primary strategies have emerged: exact algorithms\cite{lawler1966}, approximation algorithms\cite{williamson2011,vazirani2001}, and heuristic algorithms\cite{festa2014}. Exact algorithm could obtain the global optima but are prohibitive for large instances. Appoximation algorithms have theorectical gap guarantee and sometimes polynomial time complexity, but suffer from difficult construction and weak empirical performances. Heuristic algorithms are fast and ofen satisfactory in practice, but are criticised for their lack of theoretical guarantees and and need to be tailored to each specific problem. 

With the rise of deep learning in fields like computer vision and natural language processing, we now see the development of neural-based solvers for COPs. These solvers use deep learning techniques and view the COP-solving process as a learning task. The goal is to train a better heuristic using a vast number of problem instances from a specific distribution $\mathcal{D}$, such that it can effectively handle new, unseen instances from the same distribution. Once these neural models are well trained, they can be efficiently applied multiple times and solve numerous problems similar as instances from $\mathcal{D}$.

A notable contribution in the realm of neural-based solvers for COPs is S2V-DQN\cite{khalil2017}. This approach harnesses the power of graph embedding and reinforcement learning and autonomously learns efficient heuristics for various COPs. The effectiveness is proved by its performances of outperforming many expertly crafted methods on diverse problems, such as the minimum vertex cover, maxcut, and the traveling salesman problem. However, S2V-DQN has an inherent limitation. It builds a solution set by sequentially adding nodes (guided by the learned Q-value). Each node is selected only once and will not be revisited later. Consequently, this strategy yields a single one ``best guess", which, due to the high complexity of COPs, is unlikely to be the global optimum. 

An intuitive way to further improve the model's performance would be to enable the agent to keep revisiting the nodes and to add or remove them from the solution subset, until no further improvement can be made. This ever-improving adjustment can be eaisly done by heuristics like the simulated annealing\cite{kirkpatrick1983} and genetic algorithms\cite{holland1992}, which rely on sampling and iterative refinement. Yet, the same adjustment is challenging for S2V-DQN and similar finite-horizon-MDP (Markov Decision Process) based reinforcement learning(RL) models. Since such RL algorithms always start from the same configuration, and end at the same configuration, making the agent always finish the MDP within finite steps. The reason for designing MDP in this way is to help the agent learn effective heuristics more easily. Always starting from the same configuration help force the agent to learn a strategy with the same starting point, and reduce the possible trajectory space, thus requiring less training data. Finishing the MDP with finite steps forces the agent to pick the right move without allowing too many regrets. 

However, the inherent large solution space of COPs make it challenging to reach the global optima via only one single exploration. Therefore, one model that is able to explore the solution space at test time and seek ever-improving states is needed. Recently, researchers have also noticed this issue, and made some inspiring attempts. ECO-DQN\cite{barrett2020} is one such pioneering approach. However, this model was designed very laboriously, which pays great efforts on the reward design and state representations. Especially, they designed seven ad-hoc observations to represent the current state, some of which are meticulously crafted to encapsulate the long-term gain information. They proved the effectiveness of the model via empirical evaluations (implemented with the same MPNN architecture as S2V-DQN for fair comparison) on both synthetic and real-world graphs, and found that allowing the RL agent to perform reversible actions would bring lots of benefits.

In this work, we instead propose a much simpler but more effective way. Unlike ECO-DQN, we make no changes on the model architecture and training process of S2V-DQN, but only employ a technique for the RL agent during test. The technique, Gauge transformation(GT)\cite{ozeki1995,perera_computational_2020}, is very simple (less than 10 lines of Python codes implementation), but could effectively allow the trained agent reverse its earlier actions during test stage. GT achieves this by finding equivalent representations for the same problem and then resetting the current state to the same configuration as the initial one, in this way, the agent is able to continually seek improving solutions. We conduct experiments for the max-cut problem\cite{Goemans1995}, which is a canonical NP-complete combinatorial optimization problem, and has numerous real-world applications. Extensive calculations on both synthetic and real-world datasets show that GT brings great efforts over traditional RL models, like S2V-DQN. 
Furthermore, GT technique can be seamlessly integrated into various reinforcement learning frameworks, paving the way for their more effective explorations in the solving of general combinatorial optimization problems. 

The rest of this work is organized as follows. Section 2 provides an overview of the related works. Building upon this, Section 3 provides a detailed introduction to the contextual background relevant to the topic of this article. Section 4 presents a comprehensive theoretical analysis of the principles underlying GT, emphasizing its applicability within a specific scope. Furthermore, Section 5 substantiates the efficacy of GT through experimental validation. Finally, Section 6 offers a concise summary of the article and presents future perspectives for further research endeavors.
\section{Related Work}

The pioneering application of machine learning by \citet{vinyals2015} to solve COPs prompted subsequent researchers to explore various frameworks or paradigms. For instance, \citet{li2018} utilized supervised learning in conjunction with guided tree search to train Graph Convolutional Networks (GCN). Nevertheless, the supervised learning approach struggles with obtaining a substantial quantity of labels, which is especially challenging in the context of COPs due to the vast solution space. Despite these challenges, researchers persevered in exploring novel paradigms to address these issues. One such groundbreaking approach that emerged is S2V-DQN\cite{khalil2017}, which combines graph representation with RL. This innovative approach has proven to be highly effective and widely applicable in addressing COPs.

Regrettably, RL possesses an intrinsic limitation when it comes to tackling COPs. More specifically, this limitation arises from its reliance on the MDP, which imposes constraints on commencing from a predetermined state and permits each node state to be evaluated only once.This fundamental drawback poses a significant challenge for most contemporary RL-based frameworks, including S2V-DQN, when applied to COPs. Despite the continued efforts aimed at addressing this quandary, the intricate structure of the model architecture and the inclusion of intricate high-dimensional features in subsequent endeavors have a considerable impact on their generalization and accuracy.For instance, ECO-DQN\cite{barrett2020} defines actions as node flipping and employs various ad-hoc model designs such as reward shaping and input feature engineering to enhance the agent's exploration of states.
\section{Notations and Preliminaries}

A Combinatorial Optimization Problem is a type of optimization problem that seeks the extremum in discrete states:
\begin{equation}\label{eq:COPsDef}
\begin{split}
    \min_{x\in D} \quad & F(x)  \\
    \mbox{s.t.} \quad & G(x) \geq 0 \\
\end{split}
\end{equation}
Here, $x$ represents the decision variables, $F(x)$ is the objective function, $G(x)$ denotes the constraints, and $D$ represents a discrete decision space consisting of a finite set of points. 

Some COPs, such as Max-Cut problem, are defined on an inherent structure of graph $G = (V,E,W)$, where $V$ denotes the set of nodes, $E$ represents the set of edges and $W$ is the set of edge weights. Each node $v_i \in V$ and edge $e_{ij} \in E$ possess corresponding feature representations.

\subsection{Max-Cut Problem}

 The Max-Cut problem is a classic and complicated graph theory problem that is extensively researched and applied in various disciplines, including optimization\cite{elsokkary2017,venturelli2019}, computer science\cite{lu2011}, and theoretical physics\cite{barahona1982}. Moreover, it serves as a benchmark problem to assess the computational capabilities of various models and algorithms. Concretely, the Max-Cut Problem aims to partition the vertices of a graph into two separate subsets, with the goal of maximizing the cumulative weight of the edges that are severed between the subsets. Formally, Given a  graph, $G (V, E, W)$, the Max-Cut problem is to find a subset of nodes $T \subseteq V$ that maximizes the sum of edge weights in the cut set $C \subseteq E$, where the edges $(u, v)$ in $C$ satisfy $u \in T$ and $v \in V \setminus T$. The optimising function can be written as:
 \begin{equation}\label{eq:OpFunc1}
     F(u,v) = \sum_{(u,v) \in C} W(u,v)
 \end{equation}
 
\subsection{Q-learning}

Q-learning is an iterative, model-free RL method used to estimate the Q-values of state-action pairs within a MDP, denoted as $\{\mathcal{S}, \mathcal{A}, \mathcal{T}, \mathcal{R}, \gamma\}$. Here, $\mathcal{S}$ represents the state space, $\mathcal{A}$ represents the action space, $\mathcal{T}:\mathcal{S} \times \mathcal{A} \times \mathcal{S} \rightarrow [0, 1]$ is the transition function, $\mathcal{R}:\mathcal{S} \rightarrow \mathbb{R}$ is the reward function, and $\gamma \in [0, 1]$ is the discount factor. The Q-value function, which is an action value function learned by Q-learning methods, maps state-action pairs $(s,a) \in (\mathcal{S}, \mathcal{A})$ to the expected discounted sum of rewards when following a policy $\pi:\mathcal{S}\rightarrow \mathcal{A}, Q^{\pi}(s, a) = \mathbb{E}\left[\sum_{t=0}^\infty \gamma^t r_{t+1} \mid s_0 = s, a_0 = a, \pi \right]$. Compared to traditional Q-learning algorithms, DQN\cite{mnih2015} train a deep neural network $Q_{\theta}$ to approximate the value function. Hence, the policy $\pi_{\theta}$ can be determined by: $\pi_{\theta}(s) = \arg\max_{a'} Q_{\theta}(s, a')$, where $\theta$ is the set of parameters learned by DQN approach $\mathbf{N}$. Utilizing the policy $\pi_{\theta}$, one can ascertain the sub-optimal output graph $\mathbf{N}_{\pi_\theta}(G)$ , concerning the Max-Cut problem. While this outcome may accurately depict the global optimal solution in cases of smaller-scale problems, its applicability diminishes in larger-scale scenarios, necessitating an improved exploration strategy to effectively approach the attainment of global optimality.
\section{Gauge Transformation Technique}

\subsection{GT and GT foCHr graph-based COPs}
In order to utilize the technology of Gauge Transformation (GT), we need to begin by transforming the problem itself. Specifically, we need to rephrase it into an Ising formulation\cite{lucas_ising_2014}. To elucidate, we introduce the `spin’ values of {$\pm1$} for each nodes of COPs’ graph $G(V,E,W)$. Consequently, The `spin’ value embedded graph denoted as $G(V, S,E,W)$. 

Here, $S \equiv \{ \underline{s}=(s_1,s_2,\cdots,s_{|V|}) \in 2^{|V|}, s_u=\pm 1, u\in V \} $   denotes the whole `spin' configuration, $s_u$ denotes the spin value of a node. For Max-Cut problem, $s_u=+1$ if $u\in U\equiv V \setminus T$ and $s_u=-1$ if $u \in T$.

The configuration labeled as $\underline{s}_{\rm{opt}}$ corresponds to the optimal state, encompassing the ground state of the ``Ising spin glass" examined within the realm of physical sciences, or the state achieving maximum cuts in the context of the Max-Cut problem investigated in computational sciences.

Let $\underline{t} = (t_1,t_2,\cdots,t_{|V|})\in 2^{|V|},t_u=\pm 1$ be a group of so-called GT generators. A GT of `spin’ value embedded graph $G(V, S,E,W)$ can be defined as following:
\begin{equation}
    {\rm GT}(G(V,S,E,W))=G(V,{\rm GT}(S),E,{\rm GT}(W))
\end{equation}
Here,
\begin{equation}\label{eq:GTtransformation}
    \left\{
\begin{aligned}
    &    {\rm GT}(W(u,v)) = W(u,v)t_ut_v \\
    & {\rm GT}(s_u) =s_ut_u
\end{aligned}
\right.
\end{equation}

\subsubsection{Invariant property of GT.}
With the spin configuration, the optimising function (\ref{eq:OpFunc1}) can be rewritten as
\begin{equation}\label{eq:OptimalObj}
\begin{split}
    C(G)&=F(u,v)\\ 
    &= \sum_{u\in U, v\in S}W(u,v) \\
    &= \sum_{\langle u,v \rangle}W(u,v)(1-s_us_v)/2 \\
    & = -\sum_{\langle u,v \rangle}(W(u,v)/2)s_us_v + \sum_{\langle u,v \rangle}W(u,v)/2. \\
    &\equiv \sum_{\langle u,v \rangle}J(u,v)s_us_v + W'.
\end{split}
\end{equation}
The first term of (\ref{eq:OptimalObj}) is named \emph{energy function}, which is only differed in a symbol of system energy $E(G) = -\sum_{\langle u,v \rangle}J(u,v)s_us_v$ of Ising spin glass case. The second term of (\ref{eq:OptimalObj}) is a constant when the edges’ state of a graph is fixed. So that:
\begin{equation}
    C(G)=-E(G)+W'(G)
\end{equation}

The system energy is invariant under a GT.
\begin{equation}\label{eq:energyInvar}
    \begin{split}
        E({\rm GT}(G))&=-\sum_{\langle u,v \rangle}{\rm GT}(J(u,v)){\rm GT}(s_u){\rm GT}(s_v)\\
        &=-\sum_{\langle u,v \rangle}J(u,v)t_ut_vs_ut_us_vt_v\\
        &=-\sum_{\langle u,v \rangle}J(u,v)s_us_v\\
        &=E(G)\\
    \end{split}
\end{equation}

This prove is useful in expanding the exploration range of those COPs, which have the optimization function style compromised of a energy term and a constant term. 
In various other COPs, such as k-SAT/Max-SAT \cite{molnar2018continuous} and graph coloring \cite{lucas_ising_2014}, the GT approach retains the energy invariance of the system. Those COPs might contain many-body interactions, and their energy can be expressed using the following formulation:
\begin{equation}
    E(G) = -\sum_{a=1}^M J_{a}(s_{a_1}, s_{a_2}, \cdots ,s_{a_{k_a}}) s_{a_1} s_{a_2} \cdots s_{a_{k_a}}.
\end{equation}
Here the energy function $E(G)$ contains $M$ interactions, and the $a$-th interaction exists between $k_a\geq 1$ spins. In such case, property of energy invariance can be proved as (\ref{eq:energyInvar}) similarly.

\subsection{Potential reasons behind GT's success}
In this section, we explain the potential reasons why GT helps improve RL models's performances for COPs.

\subsubsection{State resetting and multiple exploration.}  
\begin{figure}[h]
    \centering
    \includegraphics[width=0.45\textwidth]{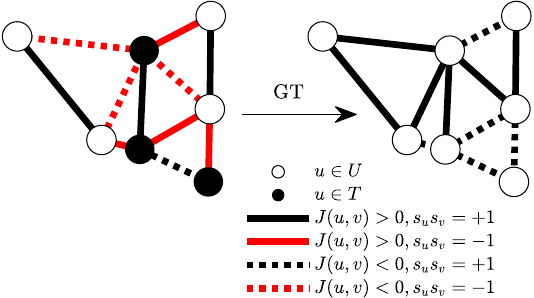}
    \caption{The illustration of GT on a graph. White and black nodes denote $u\in U\equiv V \setminus T$ and $u\in T$ respectively. Solid and dashed lines correspond to $J(u,v)>0$ and $J(u,v)<0$. $s_us_v=\pm 1$ are represented by the black/red links respectively. After GT, all $s'_u=+1$ and the weights $J(u,v)$ change their signs respectively.}
    \label{fig:GT}
\end{figure}

With the help of GT generators, graph of arbitrary spin state $\underline{s}$ can be transformed to the initial graph $G_0$, whose spin configuration is $\underline{s}_{\rm{init}}=(+1,+1,...,+1)$ , indicating $U=V,T=\emptyset$. To be specific, we just let GT generator $t_u=s_u$, so ${\rm GT}(s_u)=s_ut_u=+1$.An intuitive illustration can be seen in Fig. \ref{fig:GT}.

Without loss of generality, the approach $\mathbf{N}$ pursues the optimal result $G^{\ast}$ for the Max-Cut problem starting from the initial graph $G_0$, characterized by the node state $\underline{s}_{\rm{init}}=(+1,+1,...,+1)$. In a typical single exploration, the output is $G_1= \mathbf{N}_{\pi_\theta}(G_0)$, which might deviate significantly from $G^{\ast}$. Leveraging the property of energy invariance, the range of $\mathbf{N}$'s search expands exponentially through the integration of the GT. Moving into the next iteration of the episode, a GT is initially applied to the preceding output $G_1$. In other words, $G_2=\mathbf{{\rm GT}}(G_1)$. The exploration by $\mathbf{N}$ then uncovers $G_2’= \mathbf{N}_{\pi_\theta}(G_2)$.  Designating $\mathbf{{\rm GT}}(G_2’)$ as $G_3$, it can be proved that $G_3$ is closer to $G^{\ast}$ than $G_1$:
\begin{equation}
    \begin{split}
        C(G_3)&=-E(G_3)+W'(G_3)\\
        &=-E(G_2')+\mathbf{{\rm GT}}(W'(G_2'))\\
        & \geq -E(G_2)+\mathbf{{\rm GT}}(W'(G_2))\\
        &=-E(G_1)+\mathbf{{\rm GT}}\circ\mathbf{{\rm GT}}(W'(G_1))\\
        &=-E(G_1)+W'(G_1)\\
        &=C(G_1)
    \end{split}
\end{equation}

The second and third equations are valid by the property of energy invariance of {\rm GT}. The greater-than-equal comes from the $\mathbf{N}$'s output. The fourth equation can be easily validated by the definition (\ref{eq:GTtransformation}) of GT.

\begin{figure*}[h]
    \centering
    \includegraphics[width=0.9\textwidth]{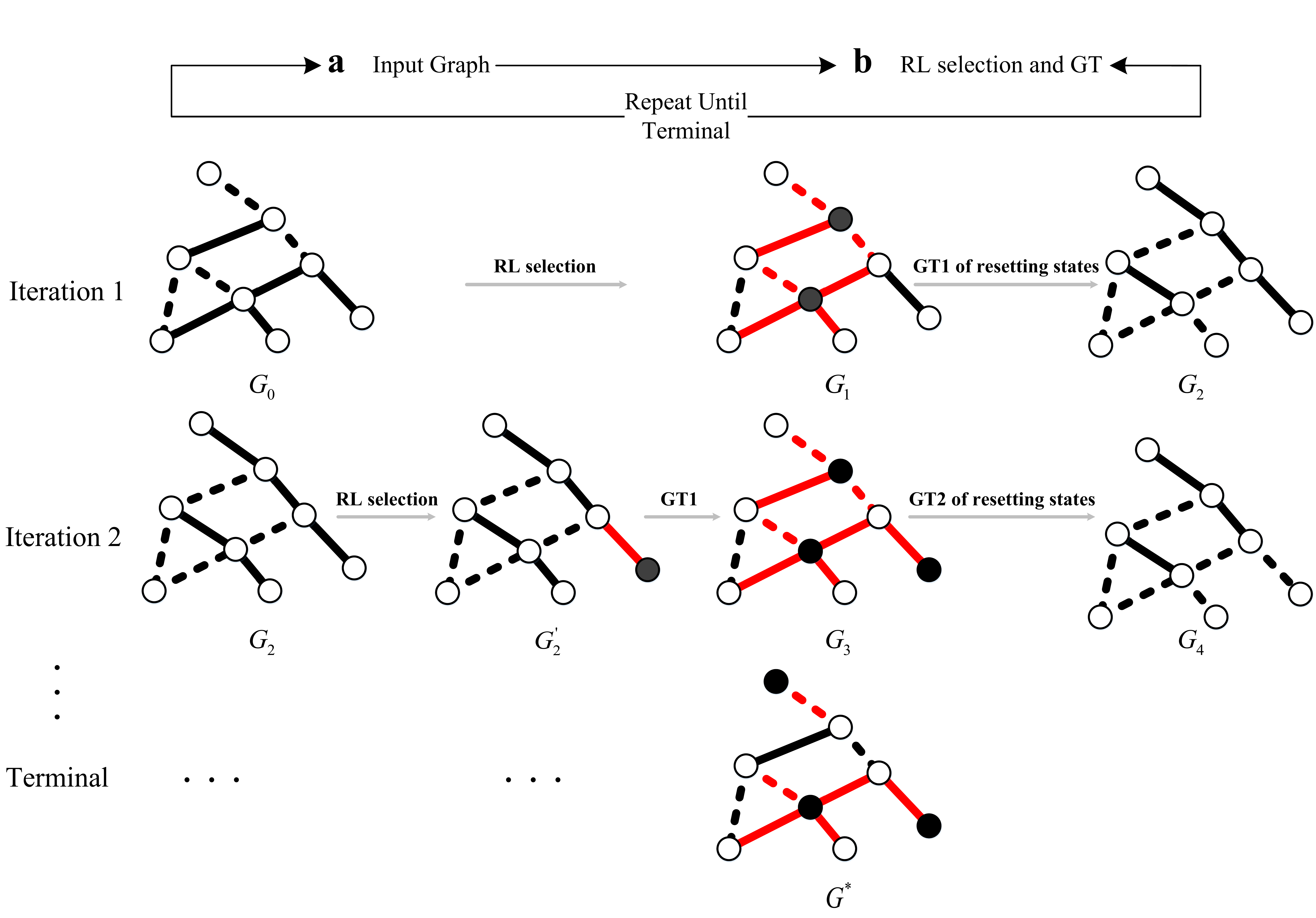}
    \caption{The illustration of GT on a graph. White and black nodes denote $u\in U\equiv V \setminus T$ and $u\in T$ respectively. Solid and dashed lines correspond to the weights of the edges are positive and negative.}
    \label{fig:GT_interations}
\end{figure*}

These iterations generate a sequence of sub-optimal graphs that progressively approximate $G^{\ast}$ more accurately: $G_1$, $G_3$, $G_5$, and so forth. By incorporating the GT into multiple explorations, the performance of approaching $\mathbf{N}$ experiences notable enhancements. Fig. \ref{fig:GT_interations} shows the details of GT.

\subsubsection{GT draw near the optimal.}
The RL-framework treats COPs as a game problem. The starting point, loss function, episode design are essential for the RL implementation for the game problem.  In simpler terms, the starting point, outcome evaluation, and action trajectory hold significant importance in effectively applying RL to tackle COPs. These lead to the aforementioned limitation of MDP. Applying $\mathbf{N}$ on a COP once, the RL agent only search through a single path in the trajectory space. However, as the complexity of the COP increases, the trajectory space rapidly expands in an exponential manner. The probability that a single path ``hits on” the global optimal state falls fast. To illustrate, consider a scenario where an optimal state within a $10\times 10$ grid network comprises 50 positive nodes and 50 negative nodes. A randomly selected path ``hits” the optimal state with a probability as slight as $\frac{1}{C(100,50)}$, less than $\frac{1}{2^{50}}$. 

The innovation brought about by GT lies in its ability to interconnect diverse exploration paths. This serves to broaden the RL agent's search scope, ultimately bolstering the likelihood of encountering the optimal state. Through GT, the RL agent's exploration process is enriched, enabling it to explore a wider array of trajectories and substantially augmenting its chances of converging towards the optimal solution.

\section{Experiments}
\begin{figure*}[htb]
    \centering
    \includegraphics[width=1\textwidth]{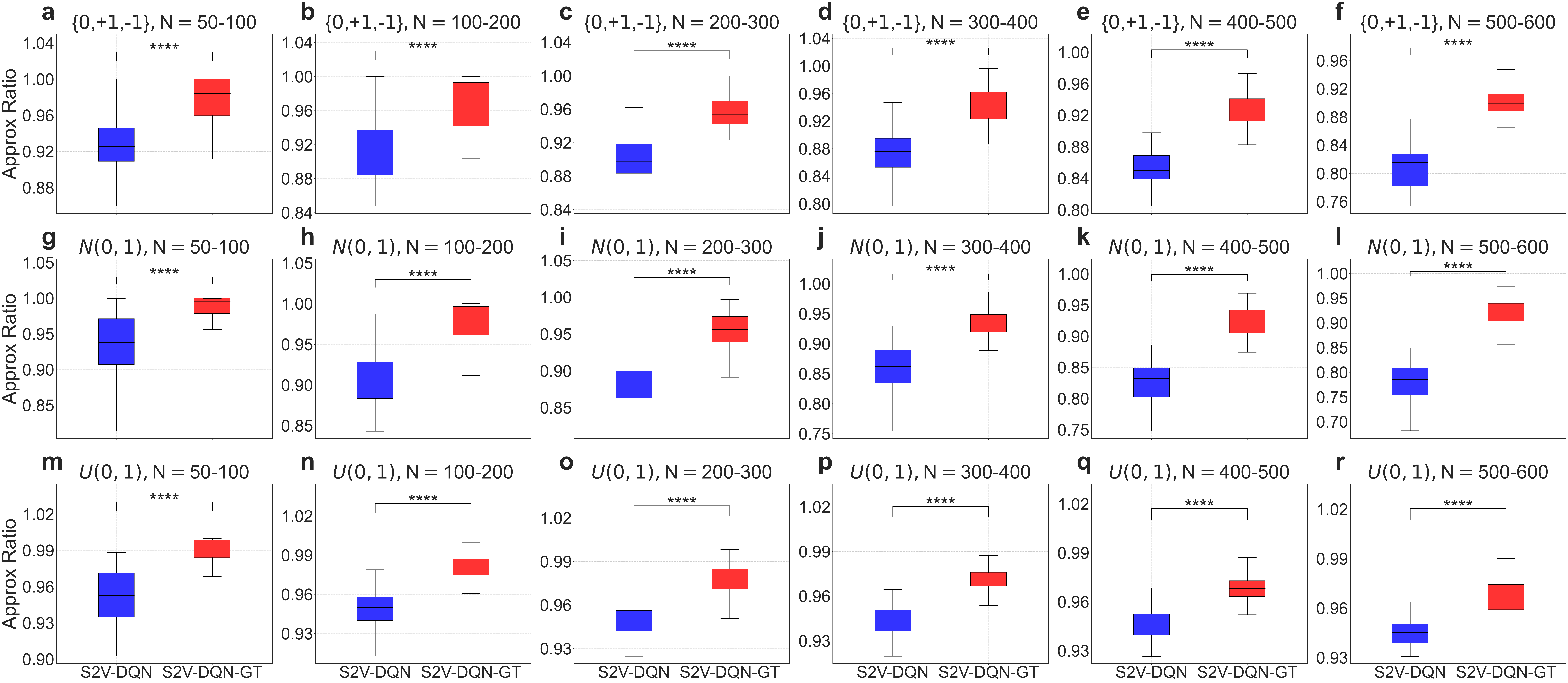}
    \caption{Statistical significance test of the performance gains brought by GT over S2V-DQN, under different distributions. We perform statistical tests between two methods (S2V-DQN and S2V-DQN-GT), so as to demonstrate the achieved gains brought by GT over traditional RL models, like S2V-DQN are not marginal, but statistically significant. For each size, we have 50 instances and report the results in a standard box-plot, together with the $p$-values of the comparisons: S2V-DQN vs S2V-DQN-GT. Here $****$ denotes $p<0.0001$. Statistical test: Wilcoxon signed-ranked test.}
    \label{fig:Pvalue}
\end{figure*}

In this section, we employ the Max-Cut problem to demonstrate the advantages of GT in addressing COPs. Our primary goal is to answer the following questions: 
\begin{itemize}
\item How many benefits would RL models obtain from GT?
\item How is GT-enhanced RL models compared with other competing methods?
\item How to utilize GT best in practice?
\end{itemize}

\subsection{Experimental Setup}

\subsubsection{Datasets.}
We conduct experiments on both synthetic datasets and real-world datasets. Synthetic graphs include Erdős-Rényi(ER) graphs\cite{erdHos1960}, Barab{\'a}si-Albert(BA) graphs\cite{albert2002}, and Watts-Strogatz(WS) graphs\cite{watts1998}. The real-world data contains Physics and Gset\cite{benlic2013}. More details can refer to the appendix A.1.

\subsubsection{Baseline methods.}
We compare our method (S2V-DQN-GT) with a wide range of competing methods for the Max-Cut problem, including greedy method, i.e. \textit{MaxcutApprox}(MCA)\cite{papadimitriou1998}, and RL-based methods, including S2V-DQN\cite{khalil2017} and ECO-DQN\cite{barrett2020}. Note that our proposed new method, S2V-DQN-GT, simply repeats several iterations of S2V-DQN, with the guide of GT, until the objective value can no longer be optimized. More details can refer to the appendix A.2.

\subsubsection{Evaluation metrics.}
To be consistent with previous work, the approximate ratio of the solution ${\rm AR}(s^{\ast})$ is defined as follow:
\begin{equation}
    {\rm AR}(s^{\ast}) = C(s^{\ast}) / C_{\rm opt}
\end{equation}
Here $C(s^{\ast})$ is the optimal solution calculated by each method, and $C_{\rm opt}$ is the optimal value, which is calculated by using the latest version of gurobi solver \cite{gurobi2022}. More details about gurobi parameters settings can refer to appendix A.3.

\subsubsection{Other settings.}
All experiments are performed on the same system with an Nvidia GeForce Tesla V100-32GB.

\subsection{Results}

\subsubsection{To what extent can RL benefit from GT?}

First of all, we only apply GT to the S2V-DQN at test time, without introducing any modifications to its underlying architecture. More details about S2V-DQN can refer to the appendix B.1. During the training phase, the model is trained on randomly generated BA graphs(average degree 4) with node number ranging from 50 to 100. Subsequently, we test the model on 50 graphs with node number ranging from 50-100 to 400-500. Ultimately, we carry out three distinct series of experiments, in which the distributions of weights assigned to the edges are respectively characterized by $\mathcal{U}(0,1)$, $\mathcal{N}(0,1)$, and $\rm{DiscreteUniform(DU)} \{0, +1, -1\}$. 

The comparison results of S2V-DQN and S2V-DQN-GT are illustrated in Fig \ref{fig:Pvalue}, and we also conduct a statistical test to show that the benefits brought by GT(i.e., S2V-DQN-GT) are not marginal. We found the achieved gains by GT are statistically significant: p-value $< 10^{-4}$ for all the cases. Based on this statistical test, we conclude that the improvement of S2V-DQN-GT over S2V-DQN is statistically significant. We also noticed that when applied to small graphs(like with 50-100 nodes), S2V-DQN-GT nearly approaches the optima. Furthermore, since all the tested model on different sizes are all trained on small scale graphs, these results also indicate the strong generalization ability of S2V-DQN-GT.

\subsubsection{How does GT-enhanced RL compare with competing methods?}

In addition, a comparison is necessary to understand how GT improves performance in relation to the current SOTA RL-based approach.To ensure a fair evaluation, each model is trained on the BA graphs containing 50-100 nodes with an average degree of 4. Notably, some methods, like S2V-DQN, are limited to starting from a specific state distribution. In order to maintain consistency, the initial node state for all methods was uniformly set to `+1’. Besides, the number of initial states of ECO-DQN is 1. More details about ECO-DQN can be refer to the appendix B.2. As a simple and efficient technique, GT can also be used in MCA, named MCA-GT. Table \ref{tab:method} summarizes the results of each method under $\mathcal{N}(0,1)$, and the results under other distributions can be found in appendix C.1, and the results on real-world datasets can be found in appendix C.2. 
\begin{table*}[htb]
  \centering
  \begin{tabular}{cccccc}
    \hline
    & 50-100          & 100-200         & 200-300  & 300-400         & 400-500         \\
    \hline
    MCA & $0.868 \pm 0.009$ & $0.850 \pm 0.006$ & $0.849 \pm 0.005$ & $0.852 \pm 0.004$ & $0.844 \pm 0.004$ \\
    S2V-DQN & \underline{0.936 ± 0.007} & \underline{0.909 ± 0.005} & \underline{0.879 ± 0.005} & $0.858 \pm 0.005$ & $0.821 \pm 0.007$ \\
    ECO-DQN & $0.928 \pm 0.006$ & $0.901 \pm 0.005$ & $0.877 \pm 0.004$ & $0.844 \pm 0.005$ & $0.800 \pm 0.007$ \\
    MCA-GT & $0.880 \pm 0.009$ & $0.859 \pm 0.006$ & $0.857 \pm 0.005$ & \underline{0.861 ± 0.004} & \underline{0.852 ± 0.004} \\
    S2V-DQN-GT & \pmb{0.987 ± 0.003} & \pmb{0.973 ± 0.004} & \pmb{0.955 ± 0.004} & \pmb{0.934 ± 0.004} & \pmb{0.925 ± 0.003} \\
    \hline
  \end{tabular}
  \caption{Comparison results of different baseline models under $\mathcal{N}(0,1)$. The best results are in bold, while the second-best ones are underlined.}
  \label{tab:method}
\end{table*}
\begin{table*}[h]
  \centering
  \resizebox{2.0\columnwidth}{!}{
  \begin{tabular}{c|c|c|c|c|c|c|c|c}
        \hline
        \diagbox{train}{test} & 15-20 & 40-50 & 50-100 & 100-200 & 200-300 & 300-400 & 400-500 & 500-600 \\
        \hline
        15-20 & \pmb{0.991 ± 0.003} & $0.984 \pm 0.002$ & $0.979 \pm 0.002$ & $0.969 \pm 0.002$ & $0.967 \pm 0.002$ & $0.960 \pm 0.002$ & $0.960 \pm 0.001$ & $0.961 \pm 0.002$ \\
        \hline
        40-50 & \textbackslash & \pmb{0.989 ± 0.002} & $0.983 \pm 0.002$ & $0.972 \pm 0.002$ & $0.967 \pm 0.002$ & $0.958 \pm 0.002$ & $0.958 \pm 0.001$ & $0.951 \pm 0.002$ \\
        \hline
        50-100 & \textbackslash & \textbackslash & \pmb{0.989 ± 0.002} & \pmb{0.980 ± 0.002} & $0.978 \pm 0.002$ & $0.970 \pm 0.001$ & $0.968 \pm 0.001$ & $0.966 \pm 0.001$ \\
        \hline
        100-200 & \textbackslash & \textbackslash & \textbackslash & $0.979 \pm 0.002$ & $0.967 \pm 0.002$ & $0.963 \pm 0.001$ & $0.956 \pm 0.002$ & $0.956 \pm 0.001$ \\
        \hline
        200-300 & \textbackslash & \textbackslash & \textbackslash & \textbackslash & \pmb{0.981 ± 0.001} & \pmb{0.979 ± 0.001} & \pmb{0.977 ± 0.002} & \pmb{0.978 ± 0.001} \\
        \hline
    \end{tabular}}
  \caption{S2V-DQN-GT's generalization in BA graphs(average degree 4).}
  \label{tab:generalization}
\end{table*}

Compared with other methods, S2V-DQN-GT achieves a notably elevated level of accuracy across all distributions. Conversely, ECO-DQN encounters challenges in sustaining accuracy when confronted with varying distributions or extensive graph sizes, due to its employment of targeted exploratory techniques to augment state exploration. Regarding MCA, GT demonstrates a certain degree of accuracy improvement, albeit not significantly pronounced. This is attributed to the fact that the selection of actions in the MCA method relies on the present state, rendering the long-term rewards of actions uncertain. Such ambiguity represents a distinctive advantage of RL-based methods.

\subsubsection{How to utilize GT best in practice?}

Further, our investigation reveals that the performance of GT is influenced by various factors, such as the characteristics and types of training graphs, the number of GT iterations, and so on. Therefore, we will analyze these factors that impact GT in the next subsection, which will facilitate practitioners in enhancing the utilization of GT.

\stitle{Training graphs.} In the training stage, the size and type of the training graph have a crucial effect on the model.

\begin{itemize}
\item \textbf{Graph types.} In the experimental setup, wherein the number of nodes ranges from 50 to 100 and the edge weight distribution follows a normal distribution $\mathcal{N}(0,1)$, three distinct graph types, namely BA, ER, and WS, are employed for separate training. The correlation heat map illustrating the results can be observed in Fig \ref{fig:differernt_types}. The findings presented in Fig \ref{fig:differernt_types} also uncover a consistent pattern observed in the field of machine learning: the predictive ability of the model on graphs from similar distributions surpasses that on graphs from dissimilar distributions.
\begin{figure}[htb]
    \centering
    \includegraphics[width=0.35\textwidth]{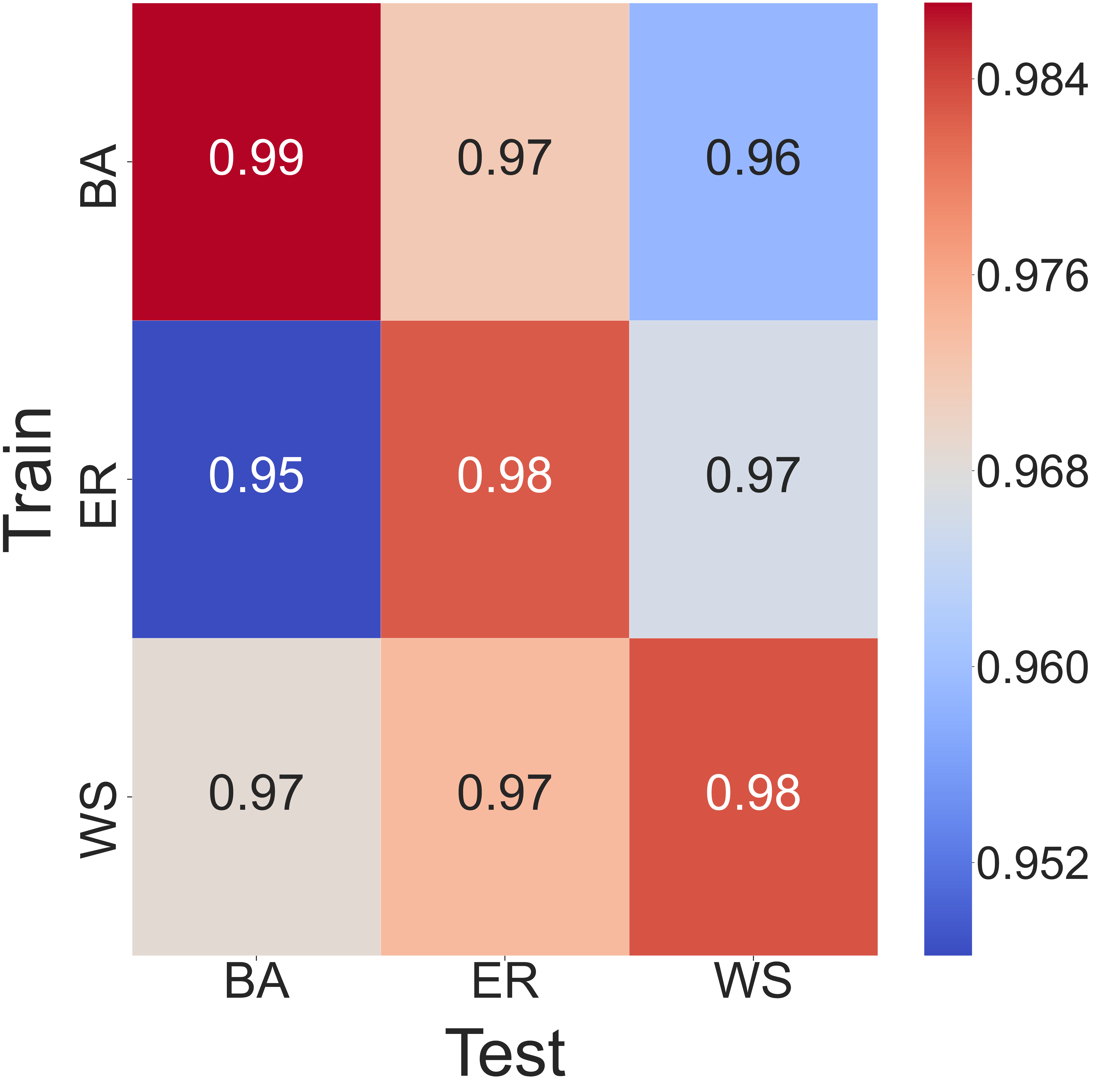}
    \caption{Heat maps under different graph types.}
    \label{fig:differernt_types}
\end{figure}
\item \textbf{Graph sizes.} In order to further elucidate the generalization capacity of GT under different graph sizes, we train the model in BA graphs under $\mathcal{U}(0,1)$. Table \ref{tab:generalization} presents compelling evidence indicating that incorporating GT substantially enhances the model's ability to generalize. Remarkably, the model exhibits consistently superior accuracy even when confronted with scenarios featuring 500-600 nodes. 
\end{itemize}

\stitle{Number of initializations.} As previously mentioned, the key advantage of GT resides in its capability to initiate exploration from distinct initial states through transformation. MCA can also be initialized from different initial states, simply by constructing the solution from a random set instead of an empty set. Importantly, we have observed that ECO-DQN incorporates techniques like Monte Carlo tree search to generate the random initial states. In order to maintain fairness, we make it a point to generate the identical random initial states for all methods. The results of these experiments are presented in Table \ref{tab:m}.
\begin{table}[h]
  \centering
  \begin{tabular}{cccc}
        \hline
        & 50-100 & 100-200  \\
        \hline
        MCA-GT (m=10) & 0.918 ± 0.006 & 0.900 ± 0.005  \\
        MCA-GT (m=100) & 0.925 ± 0.006 &  0.901 ± 0.005 \\ 
        \hline
        ECO-DQN (m=10) & 0.948 ± 0.005 & 0.921 ± 0.004 \\
        ECO-DQN (m=100) & 0.988 ± 0.002  & 0.958 ± 0.003  \\
        \hline
        S2V-DQN-GT (m=10) & 0.994 ± 0.001 & 0.980 ± 0.003  \\
        S2V-DQN-GT (m=100) & \pmb{0.995 ± 0.001} &    \pmb{0.982 ± 0.002} \\
        \hline
    \end{tabular}
  \caption{GT performance with different initial states in BA graphs(average degree 4) under $\mathcal{N}$(0,1).}
  \label{tab:m}
\end{table}

Through the results, it is evident that the utilization of a random initial states significantly enhances the accuracy of various approaches, encompassing both the greedy method and the RL-based method. Notably, the RL-based method demonstrates a notably more substantial improvement, such as S2V-DQN-GT is near optimal on the small scale graph, owing to its inclination to prioritize actions that yield greater long-term returns throughout the decision-making process. 

\begin{figure}[h]
    \centering
    \includegraphics[width=0.45\textwidth]{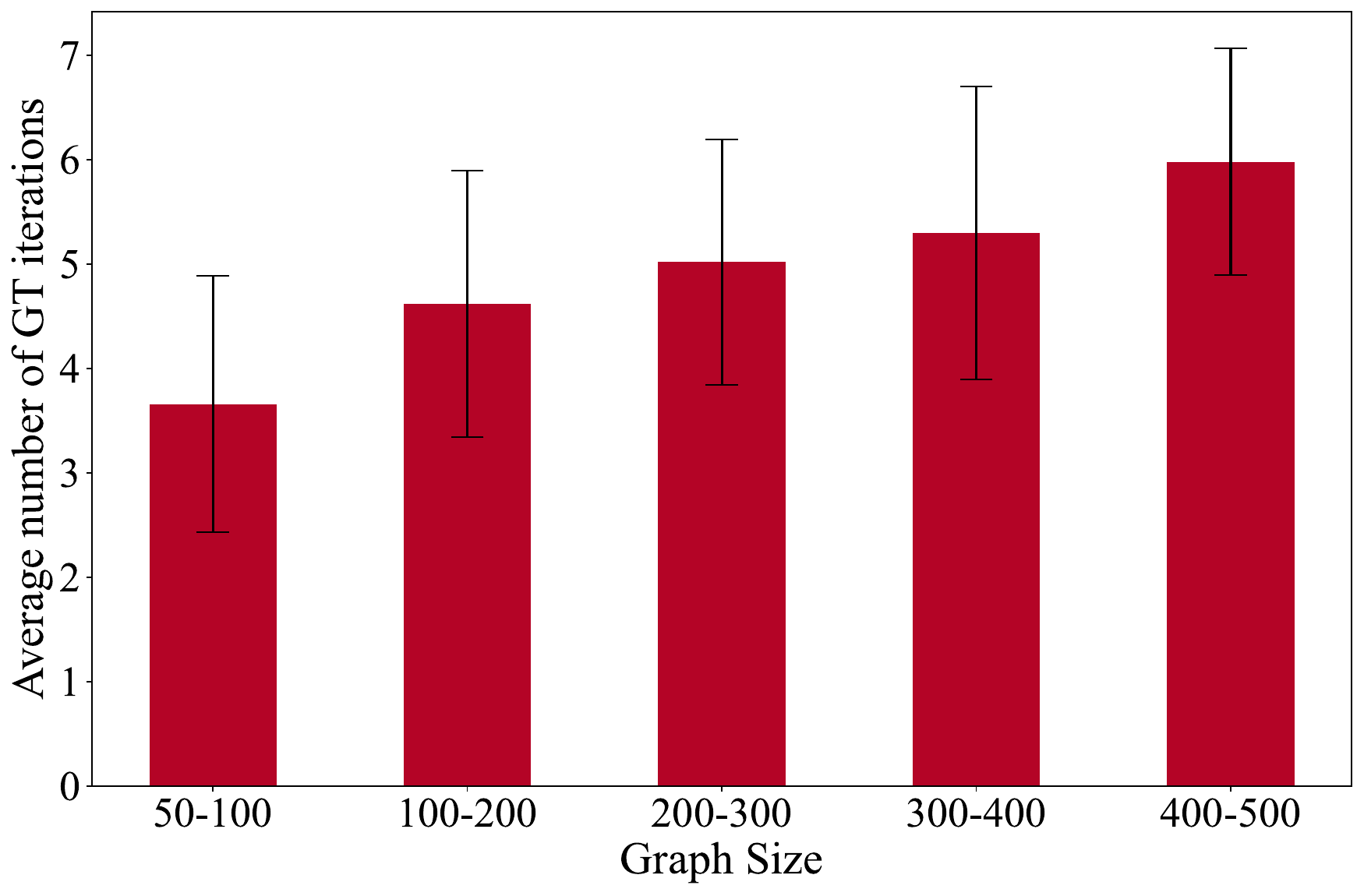}
    \caption{GT iterations with the size of BA graphs(average degree 4) with edge wights of $\mathcal{N}(0,1)$}
    \label{fig:n_trans}
\end{figure}

\stitle{Number of GT iterations.} The impact of a GT, is closely tied to the number of GT iterations. From an intuitive standpoint, a greater number of GT iterations leads to a superior outcome. Besides, as the graph's scale increases, the number of GT iterations gradually increases, as demonstrated by the specific findings presented in Fig \ref{fig:n_trans}. 

\stitle{Other discussion about GT.} Further, we found many interesting conclusions about GT through experiments, such as the GT improvement on dense graphs is better than sparse graphs. More details can refer to the appendix D.

\stitle{Suggestions for the utilization of GT in practice.} In order to better guide practitioners to use GT, we summarize the following recommendations:
\begin{itemize}
\item Training and testing graphs with identical  distributions.
\item Setting up multiple initial states can be combined with search algorithms, such as Monte Carlo tree search, to generate high-quality initial states.
\end{itemize}

\section{Conclusion}

In conclusion, our proposed method GT has successfully addressed the challenges faced by RL in tackling COPs in recent years. GT has proven to be a simple but effective technique that addresses the agent's exploration of the state, significantly enhancing the model's accuracy and demonstrating stable performance in generalization. Furthermore, we have provided a thorough theoretical analysis of GT, outlining its applicable scope and the mathematical principles behind its transformation. GT can be seamlessly integrated as a plug-in into a broader RL-based framework, opening up new avenues for solving COPs.

In the future, we can explore more powerful model architectures suitable for GT, and we can also combine with other search methods to generate high-quality initial state for GT.

\bibliography{ref}

\begin{thebibliography}{30}
\providecommand{\natexlab}[1]{#1}

\bibitem[{Albert and Barab{\'a}si(2002)}]{albert2002}
Albert, R.; and Barab{\'a}si, A.-L. 2002.
\newblock Statistical mechanics of complex networks.
\newblock \emph{Reviews of modern physics}, 74(1): 47.

\bibitem[{Barahona(1982)}]{barahona1982}
Barahona, F. 1982.
\newblock On the computational complexity of Ising spin glass models.
\newblock \emph{Journal of Physics A: Mathematical and General}, 15(10): 3241.

\bibitem[{Barrett et~al.(2020)Barrett, Clements, Foerster, and
  Lvovsky}]{barrett2020}
Barrett, T.; Clements, W.; Foerster, J.; and Lvovsky, A. 2020.
\newblock Exploratory combinatorial optimization with reinforcement learning.
\newblock In \emph{Proceedings of the AAAI conference on artificial
  intelligence}, volume~34, 3243--3250.

\bibitem[{Benlic and Hao(2013)}]{benlic2013}
Benlic, U.; and Hao, J.-K. 2013.
\newblock Breakout local search for the max-cutproblem.
\newblock \emph{Engineering Applications of Artificial Intelligence}, 26(3):
  1162--1173.

\bibitem[{Elsokkary et~al.(2017)Elsokkary, Khan, La~Torre, Humble, and
  Gottlieb}]{elsokkary2017}
Elsokkary, N.; Khan, F.~S.; La~Torre, D.; Humble, T.~S.; and Gottlieb, J. 2017.
\newblock Financial portfolio management using d-wave quantum optimizer: The
  case of abu dhabi securities exchange.
\newblock Technical report, Oak Ridge National Lab.(ORNL), Oak Ridge, TN
  (United States).

\bibitem[{Erd{\H{o}}s, R{\'e}nyi et~al.(1960)}]{erdHos1960}
Erd{\H{o}}s, P.; R{\'e}nyi, A.; et~al. 1960.
\newblock On the evolution of random graphs.
\newblock \emph{Publ. math. inst. hung. acad. sci}, 5(1): 17--60.

\bibitem[{Festa(2014)}]{festa2014}
Festa, P. 2014.
\newblock A brief introduction to exact, approximation, and heuristic
  algorithms for solving hard combinatorial optimization problems.
\newblock In \emph{2014 16th International Conference on Transparent Optical
  Networks (ICTON)}, 1--20. IEEE.

\bibitem[{Goemans and Williamson(1995)}]{Goemans1995}
Goemans, M.~X.; and Williamson, D.~P. 1995.
\newblock Improved approximation algorithms for maximum cut and satisfiability
  problems using semidefinite programming.
\newblock \emph{Journal of the ACM}, 42(6): 1115–1145.

\bibitem[{Grotschel, Junger, and Reinelt(1991)}]{grotschel1991}
Grotschel, M.; Junger, M.; and Reinelt, G. 1991.
\newblock Optimal control of plotting and drilling machines: a case study.
\newblock \emph{Mathematical Methods of Operations Research}, 35(1): 61--84.

\bibitem[{Gurobi~Optimization(2022)}]{gurobi2022}
Gurobi~Optimization, L. 2022.
\newblock Gurobi Optimizer Reference Manual (Gurobi Optimization, LLC).

\bibitem[{Holland(1992)}]{holland1992}
Holland, J.~H. 1992.
\newblock Genetic algorithms.
\newblock \emph{Scientific american}, 267(1): 66--73.

\bibitem[{KARP(1972)}]{karp1972}
KARP, R. 1972.
\newblock Reducibility among combinatorial problems.
\newblock \emph{Complexity of Computer Computations}, 85--103.

\bibitem[{Khalil et~al.(2017)Khalil, Dai, Zhang, Dilkina, and
  Song}]{khalil2017}
Khalil, E.; Dai, H.; Zhang, Y.; Dilkina, B.; and Song, L. 2017.
\newblock Learning combinatorial optimization algorithms over graphs.
\newblock \emph{Advances in neural information processing systems}, 30.

\bibitem[{Kirkpatrick, Gelatt~Jr, and Vecchi(1983)}]{kirkpatrick1983}
Kirkpatrick, S.; Gelatt~Jr, C.~D.; and Vecchi, M.~P. 1983.
\newblock Optimization by simulated annealing.
\newblock \emph{science}, 220(4598): 671--680.

\bibitem[{Laporte(1992)}]{laporte1992}
Laporte, G. 1992.
\newblock The vehicle routing problem: An overview of exact and approximate
  algorithms.
\newblock \emph{European journal of operational research}, 59(3): 345--358.

\bibitem[{Lawler and Wood(1966)}]{lawler1966}
Lawler, E.~L.; and Wood, D.~E. 1966.
\newblock Branch-and-bound methods: A survey.
\newblock \emph{Operations research}, 14(4): 699--719.

\bibitem[{Li, Chen, and Koltun(2018)}]{li2018}
Li, Z.; Chen, Q.; and Koltun, V. 2018.
\newblock Combinatorial optimization with graph convolutional networks and
  guided tree search.
\newblock \emph{Advances in neural information processing systems}, 31.

\bibitem[{L{\"u} et~al.(2011)L{\"u}, Zhang, Yeung, and Zhou}]{lu2011}
L{\"u}, L.; Zhang, Y.-C.; Yeung, C.~H.; and Zhou, T. 2011.
\newblock Leaders in social networks, the delicious case.
\newblock \emph{PloS one}, 6(6): e21202.

\bibitem[{Lucas(2014)}]{lucas_ising_2014}
Lucas, A. 2014.
\newblock Ising formulations of many {NP} problems.
\newblock \emph{Frontiers in Physics}, 2.

\bibitem[{Mnih et~al.(2015)Mnih, Kavukcuoglu, Silver, Rusu, Veness, Bellemare,
  Graves, Riedmiller, Fidjeland, Ostrovski et~al.}]{mnih2015}
Mnih, V.; Kavukcuoglu, K.; Silver, D.; Rusu, A.~A.; Veness, J.; Bellemare,
  M.~G.; Graves, A.; Riedmiller, M.; Fidjeland, A.~K.; Ostrovski, G.; et~al.
  2015.
\newblock Human-level control through deep reinforcement learning.
\newblock \emph{nature}, 518(7540): 529--533.

\bibitem[{Moln{\'a}r et~al.(2018)Moln{\'a}r, Moln{\'a}r, Varga, Toroczkai, and
  Ercsey-Ravasz}]{molnar2018continuous}
Moln{\'a}r, B.; Moln{\'a}r, F.; Varga, M.; Toroczkai, Z.; and Ercsey-Ravasz, M.
  2018.
\newblock A continuous-time MaxSAT solver with high analog performance.
\newblock \emph{Nature communications}, 9(1): 4864.

\bibitem[{Ozeki(1995)}]{ozeki1995}
Ozeki, Y. 1995.
\newblock Gauge transformation for dynamical systems of ising spin glasses.
\newblock \emph{Journal of Physics A: Mathematical and General}, 28(13): 3645.

\bibitem[{Papadimitriou and Steiglitz(1998)}]{papadimitriou1998}
Papadimitriou, C.~H.; and Steiglitz, K. 1998.
\newblock \emph{Combinatorial optimization: algorithms and complexity}.
\newblock Courier Corporation.

\bibitem[{Peng(2013)}]{peng2013}
Peng, F. 2013.
\newblock The diversified employment of China’s armed forces.
\newblock \emph{Defense White Paper}.

\bibitem[{Perera et~al.(2020)Perera, Hamze, Raymond, Weigel, and
  Katzgraber}]{perera_computational_2020}
Perera, D.; Hamze, F.; Raymond, J.; Weigel, M.; and Katzgraber, H.~G. 2020.
\newblock Computational hardness of spin-glass problems with tile-planted
  solutions.
\newblock \emph{Physical Review E}, 101(2): 023316.
\newblock Publisher: American Physical Society.

\bibitem[{Vazirani(2001)}]{vazirani2001}
Vazirani, V.~V. 2001.
\newblock \emph{Approximation algorithms}, volume~1.
\newblock Springer.

\bibitem[{Venturelli and Kondratyev(2019)}]{venturelli2019}
Venturelli, D.; and Kondratyev, A. 2019.
\newblock Reverse quantum annealing approach to portfolio optimization
  problems.
\newblock \emph{Quantum Machine Intelligence}, 1(1-2): 17--30.

\bibitem[{Vinyals, Fortunato, and Jaitly(2015)}]{vinyals2015}
Vinyals, O.; Fortunato, M.; and Jaitly, N. 2015.
\newblock Pointer networks.
\newblock \emph{Advances in neural information processing systems}, 28.

\bibitem[{Watts and Strogatz(1998)}]{watts1998}
Watts, D.~J.; and Strogatz, S.~H. 1998.
\newblock Collective dynamics of ‘small-world’networks.
\newblock \emph{nature}, 393(6684): 440--442.

\bibitem[{Williamson and Shmoys(2011)}]{williamson2011}
Williamson, D.~P.; and Shmoys, D.~B. 2011.
\newblock \emph{The design of approximation algorithms}.
\newblock Cambridge university press.

\end{thebibliography}

\clearpage
\appendix

\begin{center}
\LARGE
    \textbf{Supplementary Material}
\end{center}

\section{Detailed experimental settings}

\subsection{Descriptions of datasets} \label{sup:datasets}

\begin{itemize}
\item \textbf{Synthetic datasets.} Synthetic datasets include three well-known graphs: Erdős-Rényi(ER) graphs\cite{erdHos1960}, Barabási-Albert(BA) graphs\cite{albert2002}, and Watts-Strogatz(WS) graphs\cite{watts1998}. These graphs are commonly employed to simulate real-world complex networks. Furthermore, three diverse weight distributions are considered for the edges: the uniform distribution $\mathcal{U}(0, 1)$ within the range $[0,1]$, the normal distribution $\mathcal{N}(0, 1)$ characterized by a mean of 0 and a standard deviation of 1, and the distribution $\text{DiscreteUniform}(\rm{DU}) \{0, +1, -1\}$ where edge weights can only take the values 0, +1, or -1. Similar to previous work, the node scale of the training graph is maintained within a certain range, such as 50-100, instead of being fixed. 

\item \textbf{Real-world datasets.} We test on two real-world datasets, namely `Physics’ and Gset\cite{benlic2013} datasets. The `Physics’ encompasses a total of 10 graphs, each comprising 125 nodes and 375 edges. On the other hand, the Gset serves as a highly regarded benchmark in the field of graph research. In our work, we choose the top 10 graphs from Gset, which exhibit a substantial scale with 800 nodes. It is noteworthy that the edge weights of all graphs in these real-world datasets conform to the DU$\{0, +1, -1\}$ distribution.
\end{itemize}

\subsection{Descriptions of baseline methods}

\begin{itemize}
\item \textbf{MCA}\cite{papadimitriou1998} presents a straightforward and efficient greedy method for tackling the Max-Cut problem, wherein nodes with the largest cut changes are iteratively moved from one set to another. There exist two distinct methods for the implementation of MCA, which exhibit dissimilar approaches in constructing the solution. The first method commences the solution creation process by starting with an empty set, whereas the second method initiates the solution through a random initial set.

\item \textbf{S2V-DQN}\cite{khalil2017} blends graph representation and RL, making it the first approach to solve COPs. Its MDP possesses specific starting states exclusively, with a constraint that each state is explored only once throughout the process. 

\item \textbf{ECO-DQN}\cite{barrett2020} expands upon the S2V-DQN framework by integrating several ad-hoc designs, such as the reward-sharping, to enhance the frequency of state explorations. 

\item \textbf{S2V-DQN-GT} is to apply GT directly to S2V-DQN at test time. As shown in Fig. \ref{fig:GT_interations}, the current optimal node states found by the RL model in the last iteration are all transformed to `+1’ through GT transformation, which enables the RL model to explore once again. During each iteration of calculating the maximum cut, the state of the node undergoes the transformation again to match the real state depicted in the original graph. The termination criterion of S2V-DQN-GT is that the maximum cut value observed by RL model does not increase after applying the GT transformation.

\end{itemize}

\subsection{Gurobi parameter settings}

The latest version of the gurobi solver\cite{gurobi2022} is employed to compute the optimal value for each graph instance. The computational duration of gurobi amounts to a time span of 1 hour, during which the optimal solution is regarded as the optimal value. It is noteworthy that the slight deviation of the optimal solution obtained using the gurobi solver on large-scale graphs from the ground truth is primarily attributed to the computational time limitation. The parameter configuration employed in the computation is comprehensively outlined in Table \ref{tab:gurobi_param}. 
\begin{table}[h]
    \centering
    \begin{tabular}{cc}
        \hline
        Parameter & Value \\
        \hline
        TimeLimit & 3600 \\
        Threads & 10 \\
        MIPFocus & 1 \\
        \hline
    \end{tabular}
    \caption{Configuration of Gurobi parameters}
    \label{tab:gurobi_param}
\end{table}

\section{Hyper-parameter analysis}

\subsection{S2V-DQN's hyper-parameter analysis}






\subsubsection{Hyper-parameter configuration}

On the basis of the hyper-parameters used in S2V-DQN, we just conduct parameter tuning within the context of three distinct distributions pertaining to edge weights. Table \ref{tab:s2v_hyper} provides a comprehensive overview of the hyper-parameters that have been modified across three distinct edge weight distributions. All remaining hyper-parameters remain consistent with those employed in S2V-DQN.
\begin{table}[h]
    \centering
    \begin{tabular}{ccccc}
    \hline
         distributions   & n\_layers & n\_step & $\gamma$  \\
    \hline
        $\mathcal{U}$(0,1)   & 3 & 1 & 0.90 \\
        $\mathcal{N}$(0,1)   & 5 & 1 & 0.99 \\
        $\rm{DU}\{0, +1, -1\}$   & 5 & 1 & 0.99 \\
    \hline
    \end{tabular}
    \caption{S2V-DQN's configuration used in Experiment}
    \label{tab:s2v_hyper}
\end{table}

\subsubsection{The impact of Parameter $\gamma$ on accuracy}

In the Q-learning algorithm, the discount rate denoted by the parameter $\gamma$ determines the extent to which future rewards are taken into consideration. During the experiment, the parameter $\gamma$ is assigned a range spanning from 1 to 0.85. The results pertaining to varying values of $\gamma$ on the accuracy of the model are recorded in Table \ref{tab:hyper_gamma}.
\begin{table*}[h]
    \centering
    \begin{tabular}{cccccc}
    \hline
        $\gamma$ & 1 & 0.95 & 0.90 & 0.85 \\
    \hline
        S2V-DQN-GT & 0.864 ± 0.009 & 0.985 ± 0.002 & \textbf{0.990 ± 0.002} & 0.960 ± 0.003 \\
    \hline
    \end{tabular}
    \caption{Results of different $\gamma$ in BA graphs(average degree 4) with edge weights distribution of $\mathcal{U}(0,1)$.}
    \label{tab:hyper_gamma}
\end{table*}
\begin{table*}[h]
  \centering
  
    \text{(a) $\mathcal{U}(0,1)$}
    \vspace{5pt}
    
    \begin{tabular}{cccccc}
        \hline
        & 50-100          & 100-200         & 200-300         & 300-400         & 400-500         \\
        \hline
        MCA       & $0.949\pm0.004$   & $0.943\pm0.002$   & $0.941\pm0.002$   & $0.939\pm0.002$   & $0.941\pm0.002$   \\
        S2V-DQN   & $0.951\pm0.003$   & $0.946\pm0.002$   & $0.944\pm0.002$   & $0.940\pm0.002$   & $0.943\pm0.002$   \\
        ECO-DQN   & \underline{0.989 ± 0.002}   & \textbf{0.984 ± 0.001}   & \underline{0.970\ ± 0.001}   & \underline{0.958 ± 0.001}   & $0.933\pm0.003$  \\
        MCA-GT     & $0.958\pm0.003$   & $0.954\pm0.002$   & $0.952\pm0.002$   & $0.950\pm0.002$   & \underline{$0.950\pm0.002$}   \\
        S2V-DQN-GT   & \textbf{0.990 ± 0.002}   & \underline{0.980 ± 0.002}   & \textbf{0.978 ± 0.002}   & \textbf{0.970 ± 0.001}   & \textbf{0.968 ± 0.001}   \\
        \hline
    \end{tabular}

    \vspace{5pt}
    \text{(b) $\mathcal{N}(0,1)$}
    \vspace{5pt}
    
    \begin{tabular}{cccccc}
        \hline
        & 50-100          & 100-200         & 200-300         & 300-400         & 400-500         \\
        \hline
        MCA & $0.868 \pm 0.009$ & $0.850 \pm 0.006$ & $0.849 \pm 0.005$ & $0.852 \pm 0.004$ & $0.844 \pm 0.004$ \\
        S2V-DQN & \underline{0.936 ± 0.007} & \underline{0.909 ± 0.005} & \underline{0.879 ± 0.005} & $0.858 \pm 0.005$ & $0.821 \pm 0.007$ \\
        ECO-DQN & $0.928 \pm 0.006$ & $0.901 \pm 0.005$ & $0.877 \pm 0.004$ & $0.844 \pm 0.005$ & $0.800 \pm 0.007$ \\
        MCA-GT & $0.880 \pm 0.009$ & $0.859 \pm 0.006$ & $0.857 \pm 0.005$ & \underline{0.861 ± 0.004} & \underline{0.852 ± 0.004} \\
        S2V-DQN-GT & \textbf{0.987 ± 0.003} & \textbf{0.973 ± 0.004} & \textbf{0.955 ± 0.004} & \textbf{0.934 ± 0.004} & \textbf{0.925 ± 0.003} \\

        \hline
    \end{tabular}
    
    \vspace{5pt}
    \text{(c) $\rm{DU} \{0, +1, -1\}$}
     \vspace{5pt}
     
    \begin{tabular}{cccccc}
        \hline
        & 50-100          & 100-200         & 200-300         & 300-400         & 400-500         \\
        \hline
        MCA & $0.838 \pm 0.010$ & $0.814 \pm 0.005$ & $0.814 \pm 0.006$ & $0.803 \pm 0.005$ & $0.806 \pm 0.004$  \\
        S2V-DQN & $0.925 \pm 0.006$ & $0.912 \pm 0.006$ & \underline{0.900 ± 0.004} & \underline{0.873 ± 0.005} & \underline{0.851 ± 0.004} \\
        ECO-DQN & \underline{ 0.934 ± 0.005} & \underline{0.912 ± 0.006} & $0.890 \pm 0.004$ & $0.858 \pm 0.006$ & $0.810 \pm 0.008$  \\
        MCA-GT & $0.854 \pm 0.008$ & $0.824 \pm 0.006$ & $0.827 \pm 0.007$ & $0.812 \pm 0.005$ & $0.819 \pm 0.005$  \\
        S2V-DQN-GT & \textbf{0.976 ± 0.004} & \textbf{0.965 ± 0.004} & \textbf{0.955 ± 0.003} & \textbf{0.942 ± 0.004} & \textbf{0.925 ± 0.003}  \\
        \hline
    \end{tabular}
    
  \label{tab:all_method}
  \caption{Comparison results of different baseline models under three distributions.The best results are in bold, while the second-best ones are underlined.}
\end{table*}

\subsection{ECO-DQN's hyper-parameter analysis}

To ensure a fair evaluation of each baseline method, we introduce necessary adjustments to ECO-DQN in accordance with various experimental conditions, following the guidelines provided in the original work. 
\begin{itemize}
\item \textbf{Training graph size.} Following S2V-DQN\cite{khalil2017}, the number of nodes within the training graph is stochastically generated within a specific interval, such as 50-100, rather than being predetermined as a fixed number of all nodes present within the training graph. The fluctuations in the size of the training graph will inevitably lead to a lesser occurrence of ECO-DQN's state exploration, thereby contributing to a decrease in the precision of ECO-DQN. This outcome also reflects the limitations of some ad-hoc designs of ECO-DQN.
\item \textbf{Number of initial states.} In order to maintain consistency with the S2V-DQN, wherein the starting point is limited to a specific initial state, this paper has adopted the same constraint by setting the number of initial state states to one for all experiments, with the exception of Table \ref{tab:m}. Moreover, considering the utilization of various search techniques, including Monte Carlo tree searches, in ECO-DQN to generate superior initial states, it is noteworthy that identical random initial states are produced for all comparative baseline methods. Hence, in the experiment involving multiple initial states, we generate multiple random initial states that are identical for each graph instance. Subsequently, each method is tested on these identical random initial states individually, and then select the optimal value as the result. 
\end{itemize}

\section{Full Experimental results}

\subsection{Full results under different edge weights distributions}

Table 7 provides a full experimental results obtained for each baseline method, encompassing three distinct edge weight distributions.

\subsection{Full results on real-world datasets}

We proceed to assess the influence of GT on a practical dataset. In line with ECO-DQN, our training process focuses on an ER graph consisting of 200 nodes. The edges within this graph conform to a distribution scheme encompassing three possible values: 0, +1, and -1. Similarly, the number of initial states for all baseline methods is set to 1. Table \ref{tab:realworld} summarizes the experimental results on the real-world dataset. 

\begin{table}[h]
  \centering
  \begin{tabular}{ccc}
        \hline
          & Physics & Gset  \\
        \hline
        S2V-DQN & $0.894\pm0.012$ & $0.904\pm0.020$ \\
        ECO-DQN & $0.988\pm0.005$ & $0.983\pm0.003$  \\
        S2V-DQN-GT & $0.986\pm0.006$ & 0.978 ± 0.006  \\
        \hline
    \end{tabular}
  \caption{Comparison results of different baseline models on real-world dataset.}
  \label{tab:realworld}
\end{table}

\begin{table*}[h]
  \centering

  \vspace{5pt}
  \text{(a) BA graph}
  \vspace{5pt}
  
  \begin{tabular}{ccccc}
    \hline
    m & 4 & 6 & 8 & 10 \\
    \hline
    S2V-DQN & 0.933 ± 0.006 & 0.901 ± 0.007 & 0.708 ± 0.021 & 0.442 ± 0.023 \\
    S2V-DQN-GT & 0.989 ± 0.002 & 0.978 ± 0.004 & 0.933 ± 0.007 & 0.894 ± 0.011 \\
    performance++ & 0.057 ± 0.006 & 0.077 ± 0.007 & 0.225 ± 0.019 & 0.452 ± 0.019 \\
    GT iterations & 3.68 ± 0.144 & 3.86 ± 0.146 & 4.28 ± 0.223 & 4.30 ± 0.172 \\
    \hline
  \end{tabular}

  \vspace{5pt}
  \text{(b) ER graph}
  \vspace{5pt}

  \begin{tabular}{ccccc}
    \hline
    p & 0.15 & 0.20 & 0.25 & 0.50 \\
    \hline
    S2V-DQN & 0.913 ± 0.007 & 0.927 ± 0.006 & 0.900 ± 0.007 & 0.640 ± 0.028 \\
    S2V-DQN-GT & 0.982 ± 0.003 & 0.988 ± 0.002 & 0.979 ± 0.003 & 0.682 ± 0.035 \\
    performance++ & 0.069 ± 0.007 & 0.061 ± 0.006 & 0.079 ± 0.007 & 0.042 ± 0.020 \\
    GT iterations & 3.92 ± 0.142 & 3.76 ± 0.139 & 3.90 ± 0.167 & 4.66 ± 0.221 \\
    \hline
  \end{tabular}

  \vspace{5pt}
  \text{(c) WS graph}
  \vspace{5pt}

  \begin{tabular}{ccccc}
    \hline
    k & 8 & 10 & 15 & 20 \\
    \hline
    S2V-DQN & 0.934 ± 0.004 & 0.919 ± 0.007 & 0.566 ± 0.012 & 0.501 ± 0.015 \\
    S2V-DQN-GT & 0.982 ± 0.003 & 0.987 ± 0.002 & 0.971 ± 0.004 & 0.846 ± 0.010 \\
    performance++ & 0.047 ± 0.005 & 0.067 ± 0.007 & 0.405 ± 0.012 & 0.346 ± 0.020 \\
    GT iterations & 4.14 ± 0.183 & 4.00 ± 0.164 & 4.40 ± 0.200 & 5.74 ± 0.202 \\
    \hline
  \end{tabular}
  
  \label{tab:graph_param}
  \caption{GT performance with different graph parameters on three different types of graphs.}
\end{table*}
\begin{table*}[h]
  \centering
  \begin{tabular}{cccccc}
    \hline
    & 50-100 & 100-200 & 200-300 & 300-400 & 400-500 \\
    \hline
    MCA & 0.477 ± 0.0422 & 5.204 ± 0.447 & 25.649 ± 1.249 & 56.987 ± 2.212 & 119.193 ± 3.443  \\
    S2V-DQN & 0.888 ± 0.0446 & 3.326 ± 0.189 & 10.249 ± 0.392 & 19.018 ± 0.562 & 34.511 ± 0.981  \\
    ECO-DQN & 0.561 ± 0.0383 & 6.126 ± 0.406 & 11.344 ± 0.321 & 11.396 ± 0.316 & 11.317 ± 0.306  \\
    MCA-GT & 0.628 ± 0.056 & 6.465 ± 0.502 & 27.665 ± 1.352 & 57.860 ± 2.169 & 122.311 ± 3.434  \\
    S2V-DQN-GT & 3.881 ± 0.248 & 16.750 ± 1.083 & 49.771 ± 2.742 & 98.052 ± 5.081 & 182.352 ± 9.445  \\ 
    \hline
  \end{tabular}
  \caption{Running time of different methods. The time measurements presented are expressed in seconds (s).}
  \label{tab:time}
\end{table*}

In the real-world dataset experiment, we ensure that all experimental parameters across various methods are aligned with those of ECO-DQN. As a consequence, the parameters of the S2V-DQN-GT model are not precisely tuned to their optimal levels, leading to a slightly reduced accuracy compared to ECO-DQN. Nonetheless, it is evident that the incorporation of GT has significantly enhanced the accuracy of S2V-DQN-GT, as compared to S2V-DQN.

\section{Other discussion about GT}

\subsection{GT performance with property of graph}

Moreover, intriguing and significant findings are derived from the conducted experiment. Specifically, it is observed that as the graph instances grew in complexity, an increase in the number of GT iterations and consequently an enhancement in the model's accuracy are noted. Table 9 shows the experimental results on three different types of graphs. The model is trained in BA, ER and WS graphs with edge weights distributions of $\mathcal{N}(0,1)$.

\subsection{Trade-off between running time and approximation ratio}

Finally, a comparative analysis was performed to evaluate the computational time of different algorithms utilized on the BA graphs(average degree 4) with edge wights distribution of $\mathcal{N}(0,1)$. The specific results of this comparative analysis are provided in Table \ref{tab:time}. Notably, it is important to mention that in the comparison process, no modifications were made to the other methods themselves when applying the GT approach. As a consequence, a marginal increment in the overall execution time was observed for the broader context. However, this increment still remained in close proximity to the running time of the conventional greedy method. Remarkably, this marginal increment in time was counterbalanced by a significant enhancement in accuracy.

\section{Code and data availability}
In adherence to principles of openness and transparency, we will make all the code, data, and test graph instances associated with our research publicly accessible under an open-source license. Furthermore, upon acceptance of our work, a direct link to the open-source repository will be provided for effortless access.

\end{document}